\documentclass[runningheads]{llncs}
\usepackage{graphicx}
\usepackage{comment}
\usepackage{amsmath,amssymb} % define this before the line numbering.
\usepackage{color}

\usepackage{floatrow}

\newcommand{\ravi}[1]{{\color{red}}}
\newcommand{\lm}[1]{{\color{blue}}}
\newcommand{\yhg}[1]{{\color{magenta}}}
\newcommand{\KS}[1]{{\color{green}}}

\newcommand{\Comment}[1]{}

\newcommand{\boldstartspace}[1]{\vspace{0.1in}\noindent\textbf{#1}}
\newcommand{\shortcite}[1]{\cite{#1}}
\newcommand{\RGBA}{{RGB$\alpha$}}
\newcommand{\RGBDA}{{RGBD$\alpha$}}

\usepackage{textcomp}
\usepackage{enumitem}

\begin{document}
% \linenumbers
\pagestyle{headings}
\mainmatter
\def\ECCVSubNumber{1868}  % Insert your submission number here

\title{Deep Multi Depth Panoramas \\ for View Synthesis} % Replace with your title

\begin{comment}
\titlerunning{ECCV-20 submission ID \ECCVSubNumber} 
\authorrunning{ECCV-20 submission ID \ECCVSubNumber} 
\author{Anonymous ECCV submission}
\institute{Paper ID \ECCVSubNumber}
\end{comment}
%******************

%\begin{comment}
\titlerunning{Deep Multi Depth Panoaramas for View Synthesis}
\author{Kai-En Lin\inst{1} \and
Zexiang Xu\inst{1, 3} \and
Ben Mildenhall\inst{2} \and
Pratul P. Srinivasan\inst{2} \and \\
Yannick Hold-Geoffroy\inst{3} \and
Stephen DiVerdi\inst{3} \and
Qi Sun\inst{3} \and
Kalyan Sunkavalli\inst{3} \and
Ravi Ramamoorthi\inst{1}}
\authorrunning{K. Lin et al.}
\institute{UC San Diego \and
UC Berkeley \and
Adobe Research}
%\end{comment}
%******************
\maketitle

\begin{abstract}
We propose a learning-based approach for novel view synthesis for multi-camera 360$^\circ$ panorama capture rigs.
Previous work constructs RGBD panoramas from such data, allowing for view synthesis with small amounts of translation, but cannot handle the disocclusions and view-dependent effects that are caused by large translations.
To address this issue, we present a novel scene representation---Multi Depth Panorama (MDP)---that consists of multiple \RGBDA{} panoramas that represent both scene geometry and appearance.
We demonstrate a deep neural network-based method to reconstruct MDPs from multi-camera 360$^\circ$ images.
MDPs are more compact than previous 3D scene representations and enable high-quality, efficient new view rendering.
We demonstrate this via experiments on both synthetic and real data and comparisons with previous state-of-the-art methods spanning both learning-based approaches and classical RGBD-based methods.

\end{abstract}

\section{Introduction}
Panoramic images have been widely used to create immersive experiences in virtual environments.
Recently, commercial 360\textdegree cameras like the Yi Halo and GoPro Odyssey have made panoramic imaging practical.
However, the classical panoramic representation only allows for a 3-DoF experience with purely rotational movements; it does not support translational motion, which is necessary for a true 6-DoF immersive experience.
While recent work has leveraged panoramic depth to generate 6-DoF motion \cite{pozo2019integrated,serrano2019motion},
it is highly challenging for these RGBD-based methods to handle extensive disocclusions and view-dependent effects caused by large movements.
Meanwhile, deep learning techniques for view synthesis have demonstrated photo-realistic results \cite{mildenhall2019local,xu2019deep,zhou2018stereo};
however, these methods are not designed for panoramic inputs and rely on per-viewpoint scene representations that are expensive to store and render. %that is very expensive for high resolution panorama rendering. 

Our goal is to enable realistic, practical and efficient novel view synthesis that supports translational motion with parallax for complex real scenes.
To this end, we propose Multi Depth Panoramas (MDPs)---a novel, compact, geometry-aware panoramic representation inspired by classical layered depth images (LDIs) \cite{shade1998layered} and learning-based multiple plane images (MPIs) \cite{zhou2018stereo}.
MDPs consist of a small set of multi-layer \RGBDA{} (RGB pixel intensity, depth, opacity) panoramas that fully express complex scene geometry and appearance. 

\begin{figure}[t]
\centering  
\includegraphics[width=\textwidth]{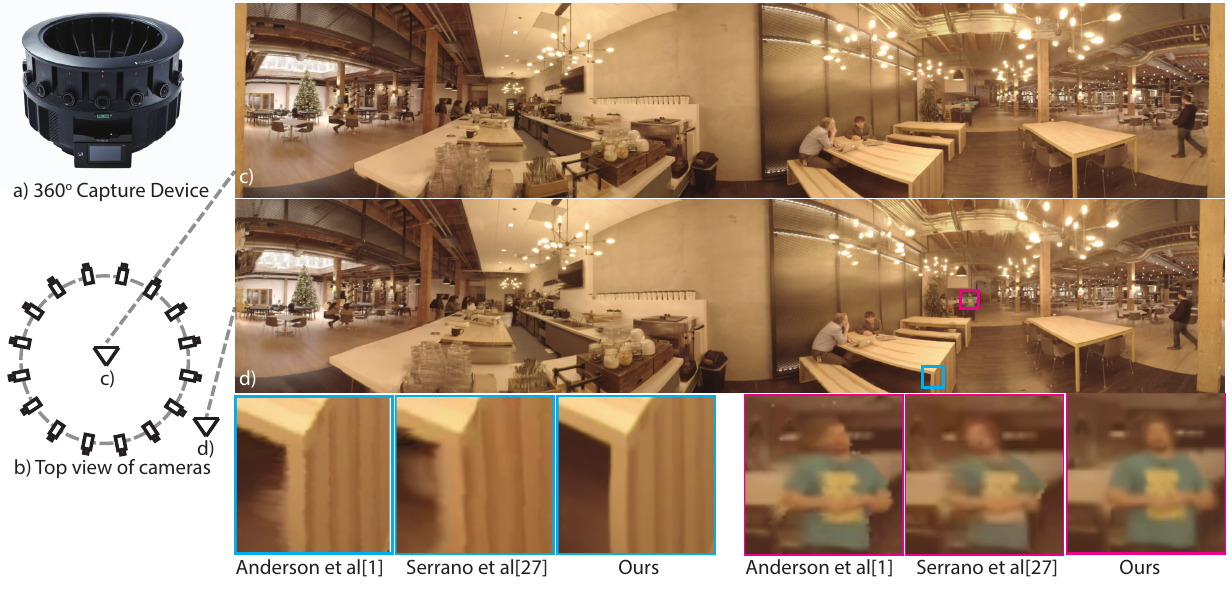}
\caption{We use 360\textdegree \ image data captured from a Yi-Halo camera (a), which consists of a ring of cameras as shown in (b). We present a learning based approach to reconstruct novel Multi Depth Panoramas (MDP) from these multi-view images, which can synthesize novel view images with both rotational and translational motions. 
We show panorama results using our MDPs from the center of the device (c) like the standard panorama synthesis, and also from a translated position (d) out of the rig. Note how the camera moves toward the counter. Our method accurately reproduces challenging disocclusion effects as shown in the cropped insets, which are significantly better than previous state-of-the-art methods that are based on RGBD representations \cite{anderson2016jump,serrano2019motion}. For better details on all the figures, please view the electronic version of this paper.}
\label{fig:teaser}
\end{figure}

We demonstrate the use of MDPs for novel view synthesis from images captured by commercial 360$^\circ$ camera rigs such as the Yi Halo that consist of a sparse array of outward-facing cameras placed in a ring.
Previous work proposes limited translational motion~\cite{serrano2019motion} using RGBD panoramas reconstructed from this setup~\cite{anderson2016jump}.
In contrast, we propose an efficient MDP-based rendering scheme that handles challenging occlusions and reflections that cannot be reproduced by state-of-the-art RGBD-based methods (see Fig.~\ref{fig:teaser}). 
Our flexible MDP representation can encode an arbitrary number of layers, degrading gracefully to a regular RGBD panorama when using a single layer. 
%In fact, each layer of the MDP representation corresponds to a RGBD panorama as used in previous work. 
MDPs are also much more compact than previous representations, providing either a similar or significantly better view synthesis quality using fewer layers, as shown in our experiments in Tab.~\ref{tab:mpi}.
Finally, by encoding the entire 360$^\circ$ panorama in a global representation, MDPs allow for complete panoramic view synthesis, unlike previous methods that focus on synthesizing limited field-of-view images (see Fig. \ref{fig:mdp}).

Our contributions can be summarized as: 
\begin{itemize}[noitemsep,topsep=0pt]
    \item a 360$^\circ$ layered scene representation called Multi Depth Panorama (MDP), offering a more compact and versatile 3D representation than previous work (sec.~\ref{sec:mdp});
    \item a learning-based method to convert images from common 360$^\circ$ camera setups into our MDP representation (sec.~\ref{sec:algo_inputs}-\ref{blending});
    \item an efficient novel view synthesis method based on MDP (sec.~\ref{splat});
    \item experiments to demonstrate the advantages of the MDP compared to existing representations (sec.~\ref{sec:experiments}).
\end{itemize}

\section{Related Work}

\boldstartspace{View synthesis.}
Novel view synthesis has been extensively studied in computer vision and graphics \cite{chen1993view},
and has been performed via approaches such as light fields \cite{gortler1996lumigraph,levoy1996light} 
and image-based rendering \cite{buehler2001unstructured,debevec1996modeling,sinha2009piecewise}.
Recently, deep learning has been applied in many view synthesis problems;
the most successful ones leverage plane sweep volumes to infer depth and 
achieve realistic geometric-aware view synthesis \cite{flynn2016deepstereo,kalantari2016learning,xu2019deep,zhou2018stereo,choi2019extreme}.
We leverage plane sweep volumes to construct per-view MPIs, similar to \cite{zhou2018stereo}.
By merging multiple MPIs from multiple views in a 360$^\circ$ camera, we construct MDPs for 6-DoF view synthesis.

\boldstartspace{Layered representation.}
Layered and volumetric representations have been applied in 3D and view synthesis applications \cite{shade1998layered,cheng2019deep,richter2018matryoshka,yao2018mvsnet,brunet2015soft,zhou2018stereo}.
Unlike a single RGBD image, 
layered representations make the scene content occluded behind the foreground viewable from side views.
In their seminal work, Shade et al.~\cite{shade1998layered} introduce the Layered Depth Images (LDIs) and Sprites with Depth (SwD) representations to render scenes using multiple RGBD images. Our methods shares the \RGBDA{} concept used in SwD, with the distinction that we reason holistically on the 360$^\circ$ scene while SwD decomposes it into multiple independent sprites. 
LDIs has been extended to a panoramic case \cite{zheng2007layered}, 
where multiple concentric RGBD panoramas are reconstructed. In addition, Zitnick et al.~\cite{zitnick2004high} utilizes multiple layers with alpha matting to produce interactive viewpoint videos. In concurrent work, Broxton et al.~\cite{broxton2020immersive} propose the multi-sphere image (MSI) representation for 6-DoF light field video synthesis, yielding accurate results but involving a memory-heavy process. In contrast with previous representations, we propose a learning-based method to reconstruct multiple memory-efficient \RGBDA{} panoramic images for 6-DoF rendering. Our method uses differentiable rendering to recover smooth object boundaries and specularities that are hard to reproduce by RGBD representations alone. 

Zhou et al. \cite{zhou2018stereo} leverage a deep network to predict multi plane images (MPIs) \cite{szeliski1999stereo} for realistic view extrapolation.
These MPIs are a dense set of fronto parallel planes with RGBA images at an input view, 
which enable rendering novel view images locally for the view.
Some recent works have extended this local view synthesis technique for large viewing ranges \cite{srinivasan19,mildenhall2019local}.
Mildenhall et al. \cite{mildenhall2019local} reconstruct MPIs at multiple views and fuse the 2D renderings 
from multiple MPIs to enable large viewpoint motion \cite{mildenhall2019local}. 
However, such a 2D image-level fusion technique requires expensively storing and rendering multiple view-dependent MPIs.
We leverage MPIs for panorama rendering and introduce multi depth panoramas (MDPs) that are view-independent \RGBDA{} images.
We propose to fuse per-view MPIs in a canonical 3D space for MDP reconstruction, which allows for efficient view synthesis from a sparse set of \RGBDA{} images.

\boldstartspace{6-DoF rendering.}
Standard 360$^\circ$ panoramic rendering only allows for 2D experience with 3-DoF rotational motion. 
Omni-directional stereo \cite{ishiguro1992omni,peleg2001omnistereo} has been introduced to provide 3D stereo
effects by leveraging panoramic depth \cite{anderson2016jump}. 
More advanced capture hardware can support 6-DoF rendering of 360$^\circ$ video~\cite{pozo2019integrated}, but there is also research to enable 6-DoF for more accessible hardware. 
Researchers have designed systems to acquire static scenes with different camera motions over time~\cite{huang2017vrvideo,luo2016parallax,bertel2019parallax,hedman2017casual} but they are unable to capture dynamic scenes without artifacts.
Other works use ring cameras to capture dynamic scenes and store the scene model in single layer panoramas~\cite{thatte2016depth} or as a panorama with a static background~\cite{serrano2019motion}, but still have significant artifacts around object boundaries, areas of poor depth reconstruction, and for large viewer head motions. 
%Many recent works have extended this to support motion parallax in full 6-DoF rendering with both rotational and translation head motion 
%to enhance the immersive experience \cite{thatte2016depth,pozo2019integrated,serrano2019motion}.
%However, such RGBD panoramic rendering tends to introduce jagged artifacts around object boundaries for large head motion. 
%We introduce multi depth panorama, a multi layer panoramic representation that expresses both foreground and background objects,
%which enables realistic 6-DoF rendering with large head motion. 

\section{Multi Depth Panoramas}
\label{sec:mdp}

\begin{figure}[t]
\centering
\includegraphics[width=\linewidth]{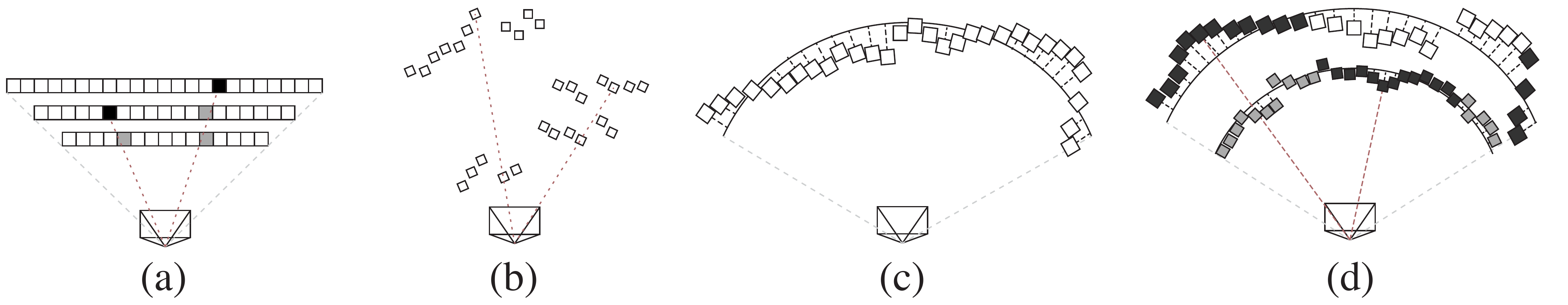}
\caption{3D scene representations: the multiplane image (a) proposes a planar representation of the scene. In contrast, the layered depth image (b) encodes multiple depth pixels along a ray.  RGBD panoramas (c) encode the scene distance per-pixel from a reference cylinder. Our multi depth panorama (d) takes inspiration from those previous representations, where depth is encoded on each cylindrical shell. }
\label{fig:mdp}
\end{figure}

\Comment{
\begin{figure}
\centering
\includegraphics[height=5.5cm]{paper_body/figures/Depth_Conflict.pdf}
\caption{TODO: add a figure to demonstrate depth conflict here. \KS{do we need this. I say we drop, especially if this is not a big issue with MPIs}}
\label{fig:depth_conflict}
\end{figure}
}
Given a 360$^\circ$ camera setup with $k$ cameras, our goal is to infer a layered scene representation that allows for high-quality 6-DoF novel view rendering. 
To achieve this, we propose the multi depth panorama (MDP) representation. %, which offers numerous advantages for novel view synthesis with 6-DoF. 
%In this section, we describe the MDP and its advantages over previous representations.

The MDP representation is inspired by Layered Depth Images (LDI)~\cite{shade1998layered} and Multiplane Images (MPI)~\cite{zhou2018stereo,srinivasan19,mildenhall2019local} (shown in Fig.~\ref{fig:mdp}(a) and (b) respectively).
LDIs and MPIs have been previously used as \emph{image-based} representations and encode only a limited field-of-view; instead, we are interested in representing the entire \emph{360$^\circ$ scene}.
Previous work has used RGBD panoramas (Fig.~\ref{fig:mdp}(c)) for novel view synthesis for panoramas~\cite{serrano2019motion} but this does not encode sufficient information to render the disocclusions that are produced at large translations. 

Motivated by these limitations of previous representations, we propose the MDP. 
An MDP consists of a sparse set of \RGBDA{} panoramas that divide the 3D space into annulus-like cylindrical regions, as shown in Fig.~\ref{fig:mdp}(d). 
%and \ref{fig:shell_collapsing}. 
Each layer $l$ encodes color $C_l$, depth $D_l$ and pixel transparency $\alpha_l$ of its corresponding region. 
Consequently, the MDP representation can be denoted by a set of \RGBDA{} layers as ${(C_0, D_0, \alpha_0), ..., (C_L, D_L, \alpha_L)}$, where $L$ specifies the maximum number of cylindrical shells.
This representation can be thought of as a set of cylindrical shells representing the 3D scene, where in each shell, a pixel is composed of five channels---diffuse RGB color, an alpha channel for occupancy, and a depth channel for finer control over 3D geometry. 
These components allow for both a more accurate and a more compact representation than previous work.
In order to synthesize novel views, we can forward splat each layer onto the target image plane and then perform the standard ``over'' alpha compositing operation \cite{porter1984compositing}.

While similar to an MPI in its use of RGB and $\alpha$ layers, our representation provides the following main benefits over the MPI:

\boldstartspace{Compactness}
The main advantage of the proposed MDP is its compactness, yielding an appreciable compression ratio over existing representations.
For example, an MPI discretizes the scene into multiple depth planes and requires a large number of such planes (32--128 \cite{zhou2018stereo,mildenhall2019local}) for high-quality view synthesis for complex scenes.
In contrast, an MDP explicitly stores the depth value allowing it to represent complex scene geometry with a small set of shells; we demonstrate this experimentally in Sec.~\ref{sec:experiments}.  
\KS{do we want to explain why we still group the depths into these shells?}
%with  for the \emph{collapse} of multiple shells into a single one without loss of accuracy (see sec.~\ref{projection} and fig.~\ref{fig:shell_collapsing}). \ravi{Why is Fig 6 referenced so many times way out of order; would it make sense to move it earlier?}
%This collapsed representation preserves the geometry accuracy and rendering quality of the original RGBA volume while using less memory as the redundant layers are combined. 

\boldstartspace{Free of depth conflicts} Our MDP representation offers a canonical scene representation for merging different views and resolving depth conflicts. 
MPI planes are created in the viewing frustum of each input camera.
In our 360$^\circ$ setup, where individual cameras are placed in an outward-facing ring, there are significant differences between these planes, making it difficult to blend between adjacent per-view MPIs.
%This issue is more severe for 360$^\circ$ cameras as individual cameras are facing outward instead of inward, amplifying this effect.
To address this issue, we adopt a canonical cylindrical 
%representation to reproject 3D points to a consistent 
coordinate system across all viewpoints,
%By choosing a unified coordinate space over a local one, our representation can 
allowing a neural network to automatically resolve depth conflicts and blend viewpoints.
This also allows us to construct a single global MDP instead of storing multiple per-view MPIs \cite{mildenhall2019local}.

\boldstartspace{Geometry accuracy} The depth component grants the MDP the same 3D representational accuracy as a point cloud. 
This increased accuracy over MPI's equidistant planes prevents depth quantization artifacts \cite{shade1998layered}, which are typically more noticeable when the scene geometry is observed from grazing angles, and leads to view synthesis results with fewer visual artifacts around geometry edges.

\section{Reconstructing and Rendering MDPs}

\begin{figure}[t]
\centering
%\vspace*{-0.35in}
\includegraphics[width=1.0\textwidth]{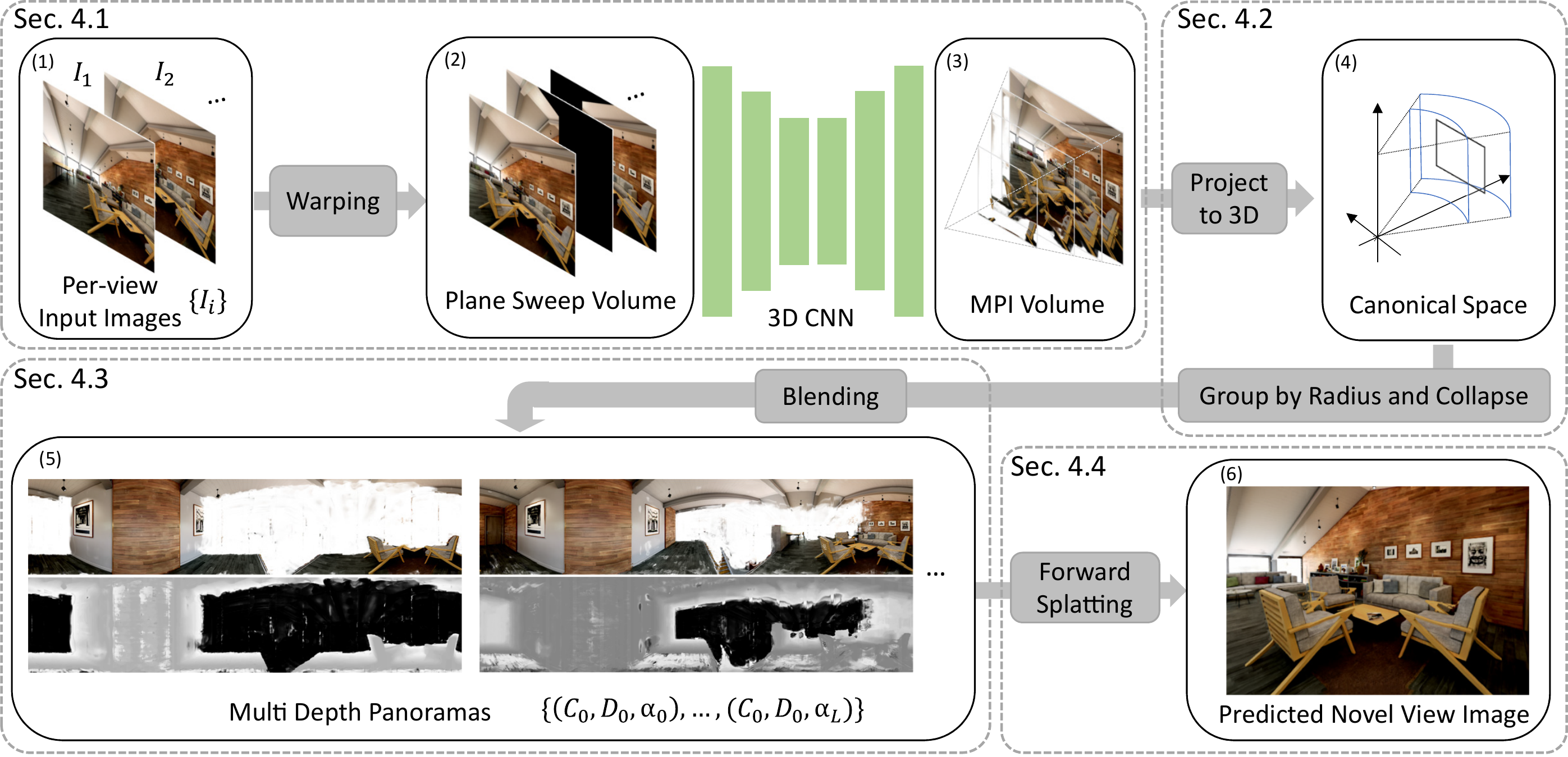}
\caption{
Overview of our pipeline.
Our network first warps multi-view images to 
each view to construct per-view PSVs, and leverages a 3D CNN to predict per-view MPIs.
These MPIs are projected to a canonical cylindrical coordinate in the world space.
We group these per-view MPIs by different radius ranges and collapse them to reconstruct per-view MDPs.
The per-view MDPs are finally blended into a single global MDP, which can be used to render new novel view images using forward splatting.}
\label{fig:algorithm}
\end{figure}

In this section, we describe our method to a) construct MDPs from images captured with a 360$^\circ$ camera setup, and b) render novel viewpoints from this representation. An overview of the full pipeline is shown in Fig.~\ref{fig:algorithm}. 

Given multiple images and their corresponding camera parameters, we first construct per-view MPI volumes (Sec.~\ref{sec:algo_inputs}). 
%To achieve this, we construct per-view plane sweep volumes similar to \cite{zhou2018stereo}. 
We then project these MPI volumes to a canonical space---a cylindrical coordinate system centered at the center of the 360$^\circ$ camera---and collapse them into a compact per-view MDP (Sec.~\ref{projection}).  
Finally, we blend the different per-view MDPs into a single MDP representation (Sec.~\ref{blending}). 
Given the reconstructed MDP representation, we can render novel views efficiently using forward splatting (Sec.~\ref{splat}).

%Next, we describe our novel layered scene representation.

% $I_0, I_1, ..., I_k$

%For each shell, we collapse the RGBA points from back to front to obtain a local displacement map as well as the accumulated RGB and alpha values (see sec. \ref{blending}).
%Finally, to render at a novel view point, we forward splat each shell onto the target image plane and alpha composite the splatted layers from back to front. (see sec. \ref{splat}).

%In the following, we first discuss our novel scene representation and its differences from previous representations.

\subsection{Reconstructing per-view MPIs from images}
\label{sec:algo_inputs}

\Comment{
\begin{figure}
  \floatbox[{\capbeside\thisfloatsetup{capbesideposition={right,top},capbesidewidth=4.5cm}}]{figure}[\linewidth]
  {\caption{We show the Yi Halo camera (left) we use for our data acquisition and the corresponding schematic camera distribution (right) from a top view. \KS{might make sense to move this to much earlier in the paper since we talk about this setup from the start.}}\label{fig:camera_config}}
  {\includegraphics[width=0.4\textwidth]{paper_body/figures/camera.pdf}}
\end{figure}
}

In the following, we describe our pipeline for predicting per-view MPI volumes from a 360$^\circ$ camera setup.
%Details for projection to a canonical space, blending and differentiable rendering are discussed in Sec. \ref{projection}, Sec. \ref{blending} and Sec. \ref{splat}, respectively.
Our pipeline takes $k$ input views and their corresponding camera parameters as input.
For each viewpoint, we create a plane-sweep volume (PSV) by warping images from the four nearest neighboring cameras as shown in Fig.~\ref{fig:algorithm}. 
%\yhg{also refer to depth conflict figure here.}
Following previous work~\cite{mildenhall2019local}, we sample the depths linearly in disparity to ensure that the resulting PSV covers accurate object depths.

\Comment{
This is for the supp mat.
\begin{figure}
\centering
\includegraphics[draft=true,height=4.5cm]{paper_body/figures/3d_cnn.pdf}
\caption{TODO: add a figure (or a table) that describes the 3D CNN network. \yhg{Due to ECCV formatting, maybe merge left/right with the previous figure?}}
\label{fig:network}
\end{figure}
}

These PSVs are processed by a 3D CNN that predicts an MPI volume.
%This MPI volume is similar to an MPI, with the exception that, in subsequent steps, we use the $\alpha$ opacity values to blend color values (as is done with MPIs), as well as depth and opacity. 
%similar to an MPI, for each PSV. \ravi{You should probably clarify differences if any.}
Please see the supplementary material for a full description of our 3D CNN. 
%This network effectively encodes depth estimation in the A channel \ravi{What exactly does this mean?}. 
Taken as is, these MPI volumes require a large amount of memory to perform novel view synthesis. 
For example, Mildenhall et al.\cite{mildenhall2019local} use 32--128 depth planes for their experiments.
%Furthermore, they have a limited range of motion before introducing visual artifacts in the synthesized view. 
We address this by projecting these MPI volumes to canonical cylindrical coordinates and collapsing them into a compact MDP representation.

%for all input volumes and each voxel is grouped into its corresponding cylindrical shell.

\subsection{Per-view MPIs to Per-view MDPs}\label{projection}

\begin{figure}
  \floatbox[{\capbeside\thisfloatsetup{capbesideposition={right,center},capbesidewidth=5cm}}]{figure}[\linewidth]
  {\caption{We show an example of MPI collapsing at a view. In particular, we show three cylindrical shells that divide the space into two bins, and five MPIs in these bins.
  We mark the MPI points that belong to different bins with different colors.
  }\label{fig:shell_collapsing}}
  {\includegraphics[width=0.5\textwidth]{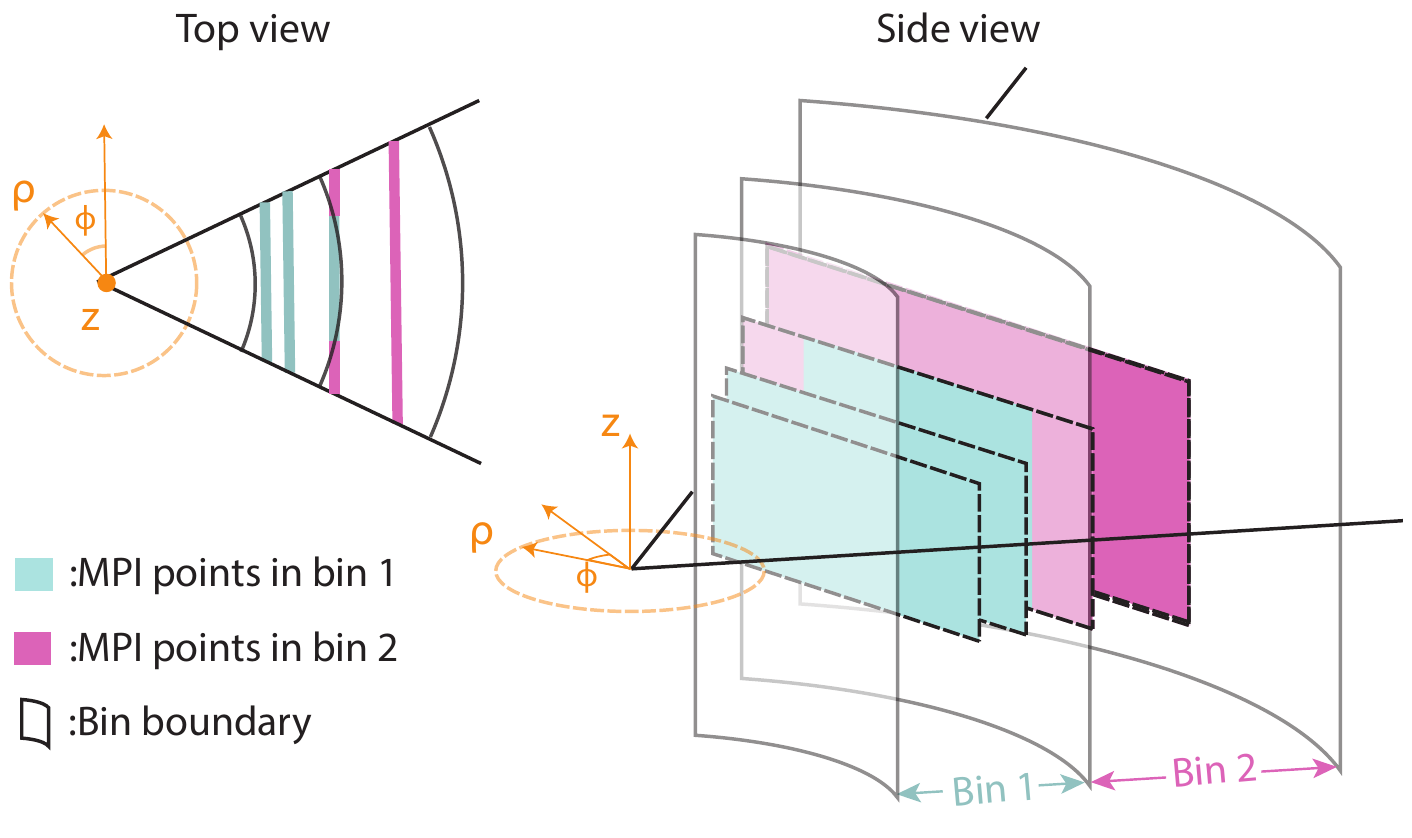}}
\end{figure}

\Comment{
\begin{figure}
\centering
\includegraphics[height=4.5cm]{paper_body/figures/collaps3.pdf}
\caption{TODO: add a figure to demonstrate shell collapsing. \yhg{be sure to name each layer \emph{l}, as noted in the text.}}
\label{fig:shell_collapsing}
\end{figure}
}
%Here, we discuss the process to project and aggregate each voxel of the RGBA volume to our cylindrical representation.
The previous step gives us, for every input view $v$, an $L$-layer MPI volume.
%an MPI volume, $M_v = {(C'_0, \alpha'_0), ..., (C'_{L'}, \alpha'_{L'})}$, where $C'_l, \alpha'_l$ denote the color and alpha channel of plane $l$, and $L$ the number of planes in the volume. 
%\ravi{Is there a reason for the primes?  Simplify notation and maybe explain briefly.  I think you can drop the primes.}
Our goal is to convert these multiple limited field-of-view volumes into a compact and unified representation. 
%Our goal is to project this limited field of view volume to a canonical space 
We do this by projecting each MPI volume to a canonical cylindrical coordinate system with its origin at the center of projection of the entire camera rig. 
Concretely, for a pixel $(x_s, y_s)$ on layer $l$ of the MPI for view $v$, its corresponding 3D point in the global world coordinates is given by:
\begin{equation}
    \begin{bmatrix}
        x_w\\
        y_w\\
        z_w\\
        1
    \end{bmatrix}
    =
    E_{w}E_{v}^{-1}I_{v}^{-1}
    \begin{bmatrix}
        x_s\\
        y_s\\
        1/d_l\\
        1
    \end{bmatrix},
\end{equation}
where $d_l$ represents the depth of layer $l$, $I_v$ and $E_v$ 
%Also, 
%\begin{equation}
%    I_s=
%    \begin{bmatrix}
%        \textbf{K}_s & 0\\
%                0 & 1
%    \end{bmatrix}\hspace{1cm}\text{and}\hspace{1cm} 
%    E_s=
%    \begin{bmatrix}
%        \textbf{R}_s & \textbf{t}_s\\
%                0 & 1
%    \end{bmatrix},
%\end{equation} 
are the intrinsic and extrinsic matrices of the camera view $v$, and $E_w$ is the extrinsic matrix for the geometric center of our camera rig.
Once projected to the world coordinates, we can calculate their cylindrical coordinates $(\rho, \phi, z)$ as: 
%treat these points as a point cloud and calculate their cylindrical coordinates $(\rho, \phi, z)$ using
\begin{equation}
    \rho_w = \sqrt{{x_w}^2 + {y_w}^2}, \;\;\; \phi_w = \arctan{(y_w/x_w)}. \label{eqn:cyl}
    %   z &= z_w.
\end{equation}
\KS{The notation was inconsistent and I have changed it. Please make sure it is correct.}

Applying the above operation to an MPI volume produces a point cloud of 3D points each with the color, depth, and opacity. 
Moreover, since each pixel coordinate in the MPI volume has multiple points at different depths, in the cylindrical coordinate system this leads to a set of multiple points along rays originating from the origin.
We collapse this large set of points into a more compact set.
More specifically, we divide the 3D space into a small number $M << L$ of annulus-like 3D cylindrical regions with equidistant radius ranges that cover the entire scene from the nearest to the farthest radius. 
We bin the 3D MPI points into these $M$ bins based on the radius $\rho_w$ (see Fig.~\ref{fig:shell_collapsing}). 
%To do this, we divide the space into a limited number $L'$ of annulus-like 3D cylindrical regions with equidistant radius ranges that cover the entire scene from the nearest radius to the farthest radius.

For each subset of points within the same bin, we use a back-to-front “over” alpha compositing operation (as is used to render novel views with MPIs \cite{zhou2018stereo,mildenhall2019local}) to compute a \emph{single} \RGBDA{} value.
Because the over operator is associative, we can process each bin individually.
%collapse each bin individually and generate the corresponding MDP layer. 
Thus, this operation replaces a large set of \RGBA{} values with a single \RGBDA{} value, thereby significantly reducing the data footprint of the representation.
In practice, as demonstrated in Tab.~\ref{tab:mpi}, we find that even 2--4 layer MDPs produce results better in visual quality than MPIs with 32 layers, leading to a significant compression.

%\emph{over} operation~\cite{10.1109/38.511,10.1145/800031.808606} to collapse the $L'$ cylindrical shells layers to the desired layer count $L$, as shown in fig.~\ref{fig:shell_collapsing}. Because the over operator is associative, we can collapse each region individually and generate the corresponding MDP layer.

% From MDP subsection, keeping it for reference:
%Our idea came from the observation that when doing over alpha-compositing operation for the MPIs, we can replace the RGB value for the depth value from each plane to obtain a depth map. Combining this with the associative property of the over operator, we can generate a partial depth map for each segment of the MPI viewing frustum and collapse the original MPI into an RGBDA representation with fewer planes.
%\ravi{It's not really clear from this, as to the key contribution of the adaptive and multi-depth.  This needs to be explained more clearly, perhaps with a figure.  This explanation seems overly complicated and hard to understand.}

\subsection{Per-view MDP Blending}\label{blending}

%\ravi{How did the primes disappear now?  This wasn't clear.}
%\ravi{Also expand this, or otherwise consider whether you need new sub-sections for text that is just one paragraph.  Also, you should reference all this to Fig. 2 clearly so readers can follow.  Notation is also ugly and flows past end of line. }

The previous step produces $k$ per-view MDPs with $M$ layers each.
%{(C^0_l, D^0_l, \alpha^0_l), ..., (C^k_l, D^k_l, \alpha^k_l)}$.
Next we blend them all into a single global MDP.
Since all the MDPs are in a canonical cylindrical representation, we can blend the individual corresponding layers. 
If layer $m$ for the view-$v$ MDP is represented as $(C^v_m, D^v_m, \alpha^v_m)$, the blended global MDP is given by:
%After creating per-view MDPs ${(C^0_l, D^0_l, \alpha^0_l), ..., (C^k_l, D^k_l, \alpha^k_l)}$ for $k$ views of a specific layer $l$, we derive a blending function $f({(C^0_l, D^0_l, \alpha^0_l), ..., (C^k_l, D^k_l, \alpha^k_l)})$ such that the output is the weighted average of all input MDPs,
\begin{equation}
    ({C}_m, {D}_m, {\alpha}_m) = \frac{\sum_v{w^v\alpha^v_m (C^v_m, D^v_m, \alpha^v_m)}}{\sum_i{w^v\alpha^v_m}}.
\end{equation}
This represents a weighted average of the color, depth, and opacity of the per-view MDPs where the weights are a product of the opacity and a per-view weight, $w^v$ that we set to the cosine of the angular difference from the optical axis, which gets lower the further a pixel is away from the principal point.

\Comment{
We treat each pixel in an MPI as a 3D point in the cylindrical space and merge these ``MPI points'' within each 3D annulus 
by splatting these points plane-by-plane into a cylindrical panorama.
This per-plane splatting is done in a far-to-near order with a novel radius-aware alpha compositing process, where both RGB values and the radii of points are accumulated with alphas across splatted planes.
This annulus-wise merging process yields partial MDPs at each view in the canonical cylindrical coordinate;
each partial MDP is an RGBDA panoramic image -- where A refers to the alpha (opacity) and D refers to the radius of the point in the cylindrical coordinate 
-- which represents the scene content seen by the view within the corresponding annulus range.
These per-view partial MDPs are then merged by pixel-wisely blending the RGBDA values across views, which constructs the full MDPs.
Here, each MDP is essentially a radius-varying cylindrical shell that can be also seen as a 3D point cloud, 
which expresses the scene geometry and appearance within the 3D annulus.
We render the MDPs by splatting the per-MDP points with alpha compositing in a far-to-near order.
}

\subsection{Differentiable MDP Rendering with Forward Splatting}\label{splat}

In this section, we describe our differentiable rendering module. 
It is achieved by forward splatting and utilizing soft z-buffering to resolve depth conflicts.

One distinction from previous MPI methods is that we cannot do planar homography warping for each layer to synthesize a novel view since each layer now lies on a cylindrical shell with a depth component. 
In order to render novel viewpoints, we treat the predicted MDP representation as a set of \RGBA{} point clouds and forward-project each point onto the target image plane.
Concretely, we can get the world coordinates $(x_w, y_w, z_w)$ by doing the inverse of the cylindrical coordinate transformation (Eqn.~\ref{eqn:cyl}) and then splat these points onto the target image plane with a bilinear kernel.

Directly projecting the points to their corresponding target location might result in depth conflicts. 
Similar to rasterization, when a query ray passes through several surfaces along its direction, the resulting pixel might simultaneously have different color information.
To ensure only the closest pixel is selected, we use z-buffering to resolve the conflicts. 
Since z-buffering is non-differentiable, we instead employ soft z-buffering~\cite{10.1007/978-3-7091-6858-5_3,lsiTulsiani18} to compute the weighted average pixel value $\bar{C}(x, y)$ of all conflicting pixels. 
Soft z-buffering can be formulated as:
\begin{equation}
    \bar{C}(x, y) = \frac{\sum C(x, y) e^{(d(x,y)-d_{max})\tau}}{\sum e^{(d(x,y)-d_{max})\tau}},
\end{equation}
where $C(x, y)$ and $d(x,y)$ denote the pixel value and inverse depth value at $(x, y)$ of a layer, respectively. $\tau$ is a scale factor to control the discriminative power of soft z-buffering. The maximum inverse depth across the image $d_{max}$ is subtracted to prevent overflow.
As a pixel gets closer, its inverse depth $d$ increases, thus increasing the weight for this pixel.
Finally, by resolving the self-occlusion depth conflicts within each layer via soft z-buffering, we can then alpha-composite the projected maps from back to front to produce the final rendering (see Fig. \ref{fig:algorithm}). \yhg{Maybe worth referring to a result figure?}\ravi{Also closely refer to parts of Fig 2 where appropriate in each subsection here to tie them together.}

\section{Implementation details}

In the following, we describe the datasets used and the training procedure to learn to project images from a 360$^\circ$ camera setup to the MDP representation.

\subsection{Dataset}

Throughout our experiments, we assume the number of cameras $k=16$, each with a 100$^\circ$ field of view, similar to commercial cameras such as the Yi Halo and GoPro Odyssey. Each camera viewing direction is 22.5$^\circ$ apart horizontally. This configuration yields a stereo overlap of over 50\% between neighboring cameras. 

In order to create photorealistic training data, we chose the Unreal Engine as our renderer since it offers complex effects and high quality rendering.
We create two datasets of different camera configurations with UnrealCV \cite{qiu2017unrealcv}: the first with a similar sampling scheme as in~ \cite{mildenhall2019local}, and the second with a ring-like camera setup similar to the Yi Halo and GoPro Odyssey.
Both datasets are generated from the same 21 large-scale scenes, which consist of indoor and outdoor scenes modeled with high resolution textures and complex scene geometry.
These datasets offer a large variety of albedo, depth complexity and non-Lambertian specular effects. 
%\yhg{We should name the datasets, so it is easier to refer to them later on, instead of "the first/second dataset."}
The first dataset contains thousands of points of interest, from each of which we sample 6 images with a random baseline to ensure various levels of disparity. Image resolution ranges from $320\times240$ to $640\times480$ to allow the network to generalize on various resolutions.
In the second dataset, the input images follow the predefined 16-camera configuration, while the output images have a random look-at direction and translational movement with a maximum radius of 25cm. Image resolution for the second dataset is adjusted to range from $320\times320$ to $512\times512$ in order to better match the field-of-view of 360$^\circ$ cameras.

\subsection{Training}

To train our method, we adopt a two-step training scheme.
During the first step, we train the 3D CNN on the first dataset to output a per-view MPI volume.
The first step is performed by selecting 6 neighboring views, using a random set of 5 as inputs and using the last image as a target view for supervision. 
During the second step, the 3D CNN is fine-tuned end-to-end with our second dataset by synthesizing a target view from its closest 5 cameras. 
%This final end-to-end step allows the 3D CNN to adjust its opacity estimation to obtain better blending. 
%This way, the network is able to generalize to the partial stereo coverage of the 360$^\circ$ camera setup. 
The network is trained using the perceptual loss from~\cite{chen2017photographic} and a learning rate of $2\times10^{-4}$ for roughly 600k and 70k iterations for each respective phase.

\section{Results}\label{sec:experiments}

\ravi{Besides the results shown here, I feel there should be qualitative results of our method at least on real scenes going back to the intro and showing the promise of full panoramas and 6DoF VR, demonstrating the actual panoramic and translational parallax qualities, which cannot be easily achieved with MPIs etc.  You may not even need comparisons, but you need to show we can do this, and maybe even allude to it in intro and teaser.  Otherwise, the current visual results, to the extent they exist at all, just look the same as any other view synthesis paper.}

We now evaluate and compare our MDP representation and panorama-based novel view synthesis method. 

\boldstartspace{Evaluation of the numbers of layers.}
Our approach works for an arbitrary numbers of MDP layers. 
In this section, we evaluate how the number of layers impact the representation accuracy using a synthetic test set that consists of two large indoor scenes different from our training set. Several hundreds viewpoints are randomly sampled within each scene. 
Table~\ref{tab:inputnumber} shows the quantitative results of our method when varying the number of layers from one to five. 
Our method consistently improves with an increasing number of layers. This increase typically allows a more accurate representation of scenes with more complex geometry and appearance.
Note that an MDP reverts to a standard RGBD panorama when its number of layers is one (see the first row of Tab.~\ref{tab:inputnumber}).
In fact, the depth obtained using an MDP with a single layer are significantly better than a state-of-the-art depth estimation method~\cite{lee2019big} (see Tab.~\ref{tab:synthetic}). \KS{can we show this visually?}
The number of layers is a user-tunable parameter that provides a trade-off between representation accuracy and memory usage. 
%Our method can be optionally used for generating this standard RGBD format, which is an additional functionality of our method; a user can decide to use either a single RGBD panorama or any number of Multi Depth Panorama layers according to their needs for quality and system resources.
We use five layers for the remaining experiments in this paper.
Note that, even with five layers, our MDPs are still compact and more memory-efficient than previous learning-based methods (see Tab.~\ref{tab:mpi}). 

\begin{table}
  \floatbox[{\capbeside\thisfloatsetup{capbesideposition={right,top},capbesidewidth=4.5cm}}]{table}[\linewidth]
  {\caption{Quantitative evaluation of the number of MDP layers. Please see the supplementary material for additional qualitative results.}\label{tab:inputnumber}}
  {\begin{tabular}{c | c c c}
        \textbf{Layers} & \textbf{PSNR}$\uparrow$ & \textbf{SSIM}$\uparrow$ & \textbf{L1}$\downarrow$\\
        \hline
        1  &        25.75  &          0.8565  &           0.0269\\
        2  &        26.17  &          0.8628  &           0.0254\\
        3  &        26.27  &          0.8655  &           0.0251\\
        4  &        26.35  &          0.8661  &           0.0251\\
        5  &\textbf{26.39} &  \textbf{0.8664} &   \textbf{0.0251}\\
    \end{tabular}}
\end{table}

\boldstartspace{Comparison with MPIs.}
Our method effectively converts the costly per-view MPIs to the novel compact view-independent MDPs.
MPIs consist of dense planes and only support local view extrapolation, which
requires rendering multiple dense sets of planes from multiple views to enable arbitrary rotational motion. 
In contrast, our MDPs are in a canonical world space, 
which only requires splatting a single sparse set of depth layers for 360\textdegree \ 6-DoF view synthesis.
In Tab.~\ref{tab:mpi}, we show the quantitative results of our method with 2 and 5 layer MDPs.
We also compare against a naive solution that directly uses the per-view MPIs to do view synthesis with 16-view MPIs and 32 planes per MPI.
For this, following the method of Mildenhall et al. \cite{mildenhall2019local}, we linearly blend the five neighboring per-view MPI renderings with the pixel-wise cosines of the angles between the per-pixel viewing directions and the central direction of each camera.
The corresponding memory usage of these MPIs and MDPs are shown in the table.
%The accuracy of our MDP representation outperforms the MPI. 
The naive MPI method is not able to effectively blend the multi-view renderings; training it end-to-end to learn the 2D blending process may improve the quality.
In contrast, our approach can leverage its priors learned during training to merge the per-view MPIs, which leads to a method that outperforms the MPI approach. 
Moreover, our method is significantly more memory-efficient, taking an order of magnitude less memory than the MPI method. 
%This demonstrates the compactness of our MDP, which potentially enables applications in low-memory mobile devices.

\begin{table}[h!]
  \begin{center}
    \caption{Quantitative results of our method compared to a MPI-based method on synthetic scenes, along with their memory usages. Please see the supplementary material for more qualitative results.}
    \label{tab:mpi}
    \setlength{\tabcolsep}{5pt} % Default value: 6pt
    \begin{tabular}{c | c c c | c c}
        \textbf{Methods} & \textbf{PSNR}$\uparrow$ & \textbf{SSIM}$\uparrow$ &\textbf{L1}$\downarrow$ & \textbf{Dimension}     & \textbf{Storage} \\
        \hline
        MPI              &24.75  &  0.8278  &   0.0306 & $16 \times 640 \times 640 \times 32 \times 4$ & 3.355GB \\
        Ours - 2 Layers  &26.17  &  0.8628  &   0.0254 & $2560 \times 640 \times 2 \times 5$ & \textbf{0.066GB} \\
        Ours - 5  Layers  &\textbf{26.39}  &  \textbf{0.8664}  &   \textbf{0.0251} & $2560 \times 640 \times 5 \times 5$ & 0.165GB \\
    \end{tabular}
  \end{center}
\end{table}

\boldstartspace{Comparisons with other methods on synthetic scenes.}
We now compare our method with other 360\textdegree \ view synthesis techniques that allow for translational motion and demonstrate quantitative comparisons on our synthetic testing set.
The most popular way to do 360$^\circ$ rendering is to reconstruct the depth of a panorama 
and render the RGBD panorama as a point cloud or mesh, as introduced in \cite{pozo2019integrated,anderson2016jump}.
These techniques rely on a pre-computed depth map, 
and we use a state-of-the-art deep learning based depth estimation method \cite{lee2019big} 
to generate depth for input panoramas.
This depth map is used to generate a corresponding RGBD point cloud and mesh that are used in turn to render novel views.
We also compare against a state-of-the-art method \cite{serrano2019motion}
that is designed to improve the 2D images rendered from the RGBD panorama mesh 
by resolving the disocclusions around the geometric boundaries.
%\ravi{It may be worth making our results (best) bold in all these tables to highlight them.}

\begin{table}[t]
  \floatbox[{\capbeside\thisfloatsetup{capbesideposition={right,top},capbesidewidth=5cm}}]{table}[\FBwidth]
  {\caption{Comparison on synthetic scenes. We compare with other methods that use a single RGBD panorama for 6DoF rendering.
  We show results of SSIMs, PSNRs and L1 loss of these methods.}\label{tab:synthetic}}
  {\begin{tabular}{c | c c c c c}
    \textbf{Methods} & \textbf{PSNR}$\uparrow$ & \textbf{SSIM}$\uparrow$ & \textbf{L1}$\downarrow$\\
    \hline
    Depth \cite{lee2019big} as points           &      23.06  &  0.766  &   0.047\\
    Depth \cite{lee2019big} as mesh             &      23.75  &  0.780  &   0.040\\
    Depth \cite{lee2019big} + \cite{serrano2019motion}       &      23.20  &  0.767  &   0.043\\
    Our depth + \cite{serrano2019motion}               &      24.98  &  0.827  &   0.031\\
    Our single layer MDP                                     &      25.75  &  0.857  &   0.027\\
    Our MDPs (5 layers)                                &      \textbf{26.39}  &  \textbf{0.866}  &   \textbf{0.025}\\
    
    \hline
  \end{tabular}}
\end{table}

\begin{table}[h!]
  \begin{center}
    \caption{Quantitative comparison on maximum disparity. Our method consistently outperforms~\cite{serrano2019motion} on image quality over all tested disparity levels.}
    \label{tab:disp}
    \setlength{\tabcolsep}{7pt}
    \begin{tabular}{c | c c c}
        \textbf{Disparity (PSNR/SSIM)} & \textbf{32} & \textbf{64} &\textbf{128} \\
        \hline
        Depth \cite{lee2019big} + \cite{serrano2019motion} & 22.40 / 0.7490  &  20.20 / 0.6911  &   18.17 / 0.6322 	 \\
        Our MDPs (5 layers)  & 26.90 / 0.8839 &  24.96 / 0.8436  & 21.94 / 0.7579
    \end{tabular}
  \end{center}
\end{table}

\begin{figure}[t]
    \includegraphics[width=1.0\textwidth]{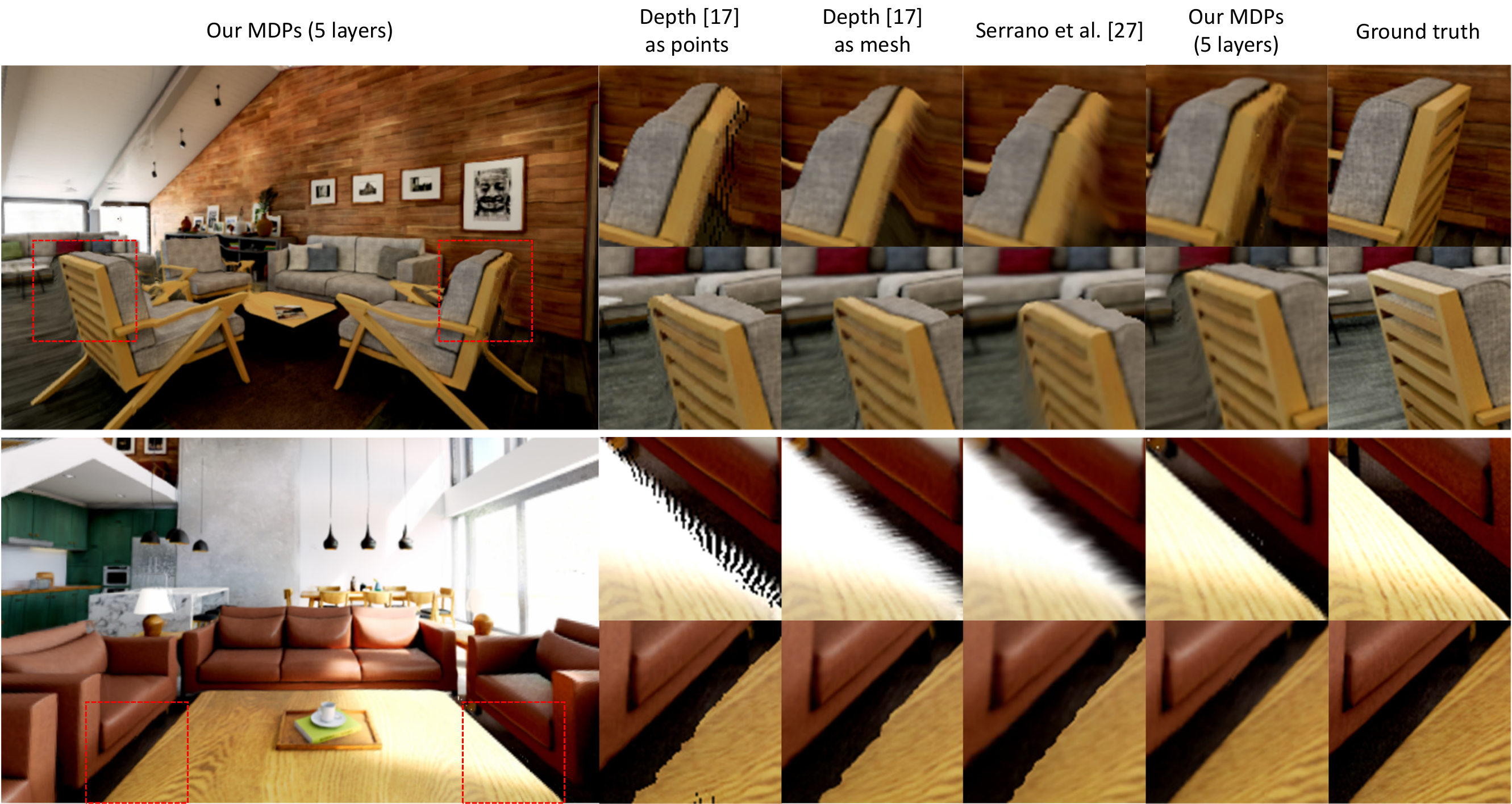}
    \caption{Visualization of synthetic results. We show two examples of synthetic results to illustrate the visual quality corresponding to the numbers in Tab.~\ref{tab:synthetic}.}
    \label{fig:synthetic}%
\end{figure}

Table~\ref{tab:synthetic} shows the quantitative results of these methods.
Our MDPs generate significantly better results than the other comparison methods as reflected by the highest PSNRs and SSIMs and the lowest L1 loss.
We also show our single layer MDP result in Table~\ref{tab:synthetic},
which is essentially doing single RGBD panorama-based point cloud rendering. 
Interestingly, even this performs better than the point cloud rendering using the prior state-of-the-art depth estimation method.
This demonstrates that our method can be used as an effective panorama depth estimation technique by estimating a single MDP.
We also show that, by giving the depth from our single MDP as input to Serrano et al.'s enhancing technique~\cite{serrano2019motion}, it improves their result over using the depth from Lee et al.~\cite{lee2019big}. 
Note that, Serrano et al.~\cite{serrano2019motion} focus on improving user experience and mitigate unpleasant artifacts, which may decrease the accuracy of direct mesh rendering. 
% \KS{what is this referring to?}

To visualize the comparisons, we show two results on synthetic data in Fig.~\ref{fig:synthetic}.
Our results offers increased visual quality than all other comparison methods, especially around object boundaries.
Our results are also the most similar to the ground truth images,
which is consistent with the quantitative results in Tab.~\ref{tab:synthetic}.
Note that, a single RGBD representation used in other methods is very limited 
and cannot well reproduce the challenging disocclusion effects.
This leads to holes in the point cloud rendering, and noisy and stretched boundaries in the mesh rendering; 
Serrano et al. \cite{serrano2019motion} smooth out the boundaries in the mesh rendering, which in turn introduces blurriness and distortion.
Our MDPs can describe complex scenes with multiple objects on the line-of-sight and effectively handle challenging boundaries, disocclusions and other appearance effects.

We now analyze the robustness of our method on large disparities. To do so, we translate the target viewpoints in our test set and record the amount of maximum scene disparity with respect to the reference viewpoint. We use a 32-layers MPI to perform the sampling. We compare our method to~\cite{serrano2019motion} and report the results in Tab.~\ref{tab:disp}.
%We report this for selected disparity levels (32-128).
%and overall mean for all disparity levels (4-128). \yhg{where?}
In this table, we focus on larger disparity levels as these correspond to larger translations. As expected, larger translations yields a more difficult task, which reduces the novel view visual quality. Despite this, our method outperforms~\cite{serrano2019motion} across all disparity levels. These observations also match the sampling guidelines in Mildenhall et al.~\cite{mildenhall2019local}.

\begin{figure}[t]
    \includegraphics[width=0.95\textwidth]{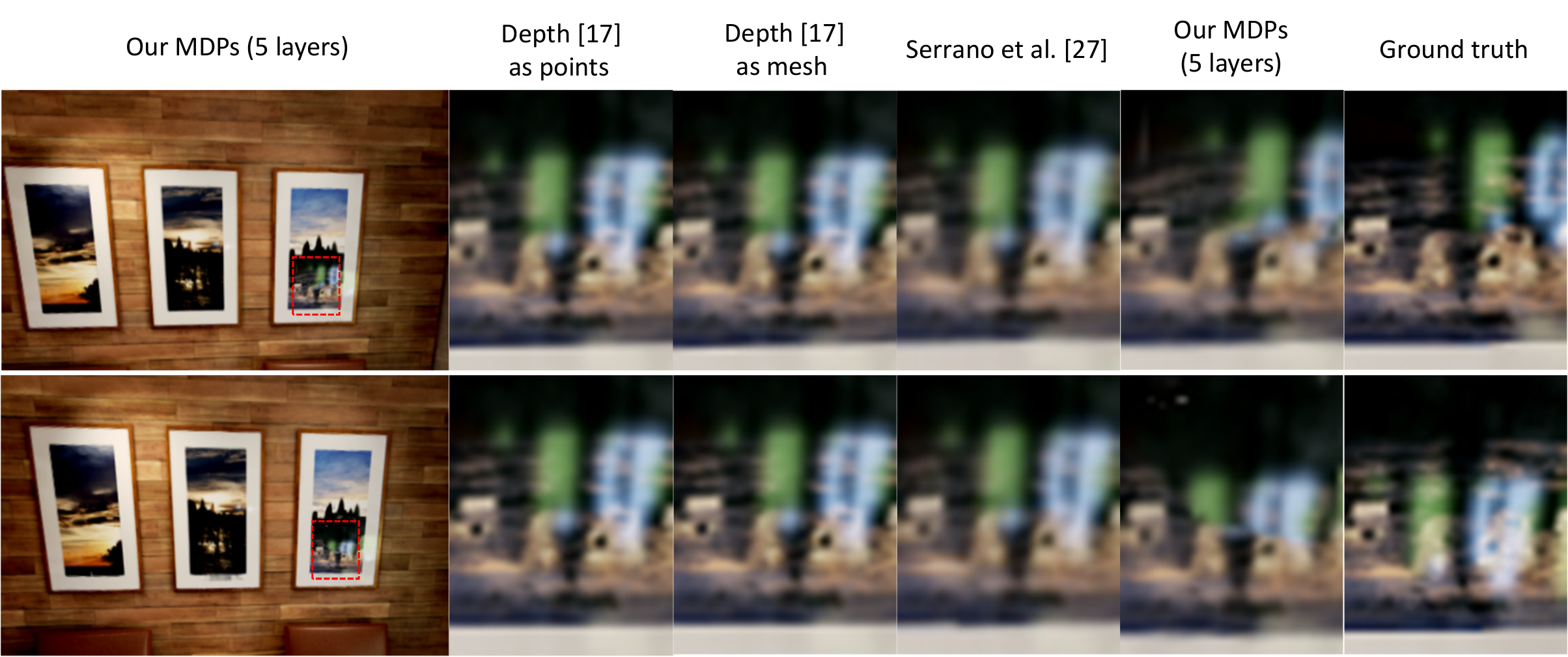}
    \caption{Reflection effects. We show one example of specular reflection with zoomed-in insets. Note that, our method recovers the specular reflection motion that matches the ground truth very well, while the reflections in the comparison methods do not move.}
    \label{fig:specularity}%
\end{figure}

We also show an example of specular reflection in Fig.~\ref{fig:specularity}.
Note that the other methods use a single depth, which bakes the reflections in the object color and fails to accurately capture the motion of reflections from different viewpoints. 
In contrast, our novel MDP allows the reflections to be encoded at a different depth, which effectively models the moving reflection.

 %leverage a single depth map, which bake the reflecting and reflected objects together in the colors and cannot express moving reflections. In contrast, our MDPs allows the reflections to be located at different depth; our method effectively models the moving reflection with changing viewpoint moves, which matches the ground truth very well.

\begin{figure}[t]
    \centering
    %\vspace*{-0.5in}
    \includegraphics[width=\textwidth]{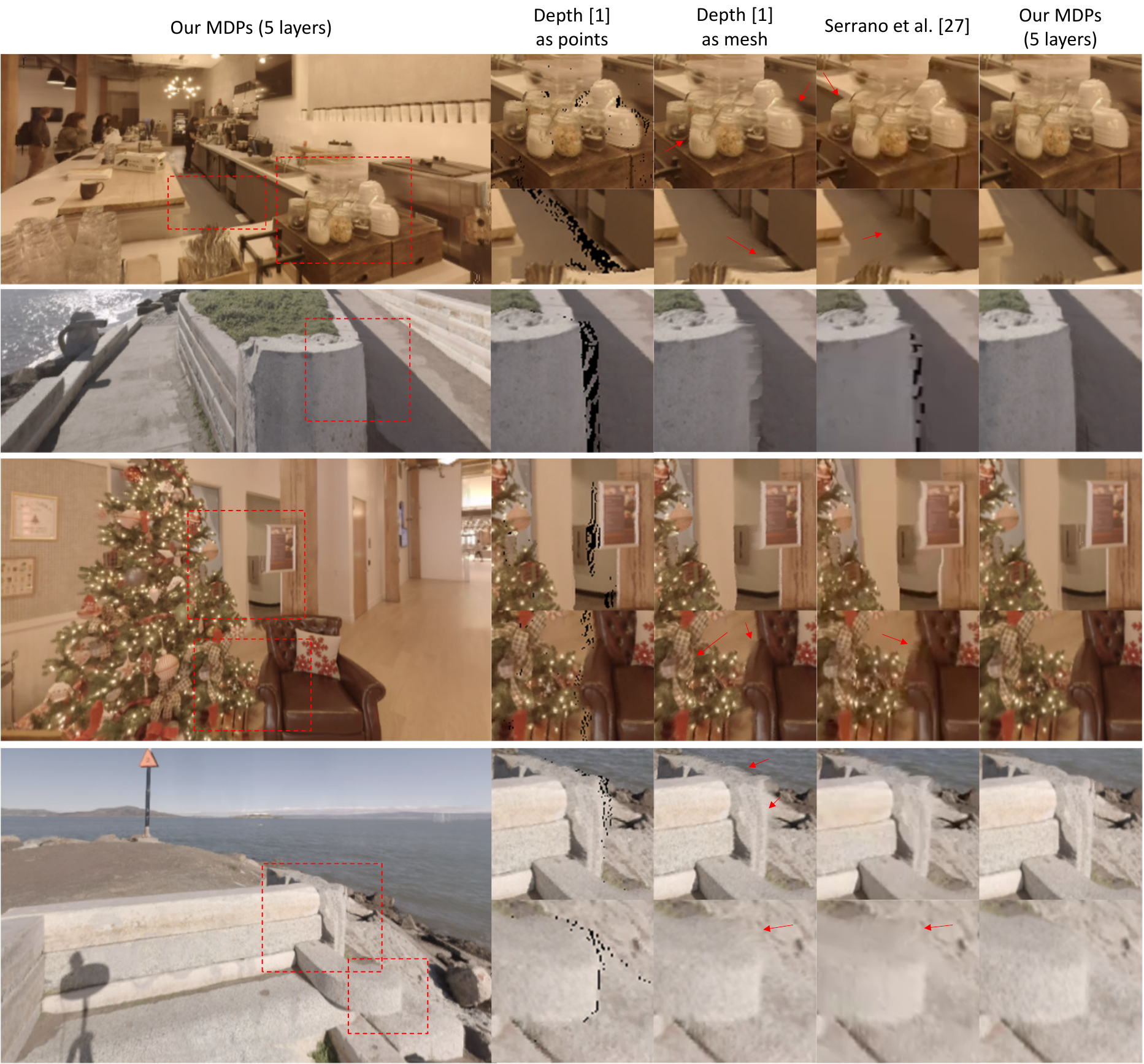}
    %\vspace*{-2.0in}
    \caption{Qualitative results on complex real scenes. We show results on novel viewpoints comparing our 5-layers MDP (leftmost) to RGBD based methods. }
    \label{fig:real}
\end{figure}

\boldstartspace{Comparisons on real scenes.}
We now evaluate our method on complex real scenes and compare them with 
the methods using RGBD panoramas.
We captured these real scenes with a Yi Halo and used Google Jump Manager \cite{anderson2016jump} to stitch the multi-view images and generate the depth. 
%which is specifically designed for panorama synthesis on real data.
Figure~\ref{fig:real} shows the results of our method with 5-layers MDPs, rendering an RGBD panorama as a point cloud and mesh, 
and using \cite{serrano2019motion} to improve the mesh rendering.
Similar to the synthetic comparisons, our method produces more realistic results
than the comparison ones.
Note that, the disocclusion effects introduce obvious holes and significant discontinuous noise 
in the baseline point cloud rendering and mesh rendering techniques.
While \cite{serrano2019motion} resolves the noisy boundaries in the mesh rendering 
for better user experience, 
the edges are in fact distorted in many regions and still look physically implausible. 
Our results are significantly more photorealistic. Please refer to supplementary material for more experiments and ablation study on end-to-end training.

%\KS{\boldstartspace{Limitations}}

\boldstartspace{Limitations}
Our proposed MDP representation can be used to represent complex geometry and appearance, given enough layers.
%This assumes the multiple layers of MDPs fully depict the multiple modalities of the scene geometry along each panoramic pixel ray.
While five layers are sufficient for most cases as previously shown, it might not be adapted to more complex and challenging scenes and can exhibit blurriness or distortion. Increasing the number of layers can potentially address this. 
Besides, our novel view synthesis method is limited to relatively small motions for two reasons. First, large translations will potentially expose large unseen parts of the scene, not encoded in the MDP representation. We hypothesize that future work on generative models might provide a solution to this issue. Second, moving past the first concentric sphere of the MDP would break the alpha ordering.
\section{Conclusion}

We presented a novel 3D scene representation---Multi Depth Panoramas, or MDP---that represents complex scenes using multiple layers of concentric \RGBDA{} panoramas. 
MDPs are more compact than prior scene representations such as MPIs. As such, they can be used to generate high-quality novel view synthesis results with translational motion parallax, using our proposed forward-splatting rendering method. 
Furthermore, we presented a learning-based method to accurately reconstruct MDPs from commercial 360$^{\circ}$ camera rigs.

\boldstartspace{Acknowledgements}
We would like to thank In-Kyu Park for helpful discussion and comments. This work was supported in part by ONR grants N000141712687, N000141912293, N000142012529, NSF grant 1617234, Adobe, the Ronald L. Graham Chair and the UC San Diego Center for Visual Computing.

%While we have explored the use of MDPs for image view synthesis, their compact nature is especially well suited for panoramic video applications, where the large memory footprint of an MPI is prohibitively expensive.
%Extending MDPs to this scenario, and especially handling static background and dynamic foreground elements would be interesting directions to explore. 

\clearpage
\bibliographystyle{splncs04}
\bibliography{egbib}
\end{document}